\documentclass[a4paper,fleqn]{cas-sc}

\usepackage[authoryear]{natbib}

\def\tsc#1{\csdef{#1}{\textsc{\lowercase{#1}}\xspace}}
\tsc{WGM}
\tsc{QE}

\begin{document}
\let\WriteBookmarks\relax
\def\floatpagepagefraction{1}
\def\textpagefraction{.001}

\shorttitle{}    

\shortauthors{}  

\title [mode = title]{Dual-Prompt CLIP with Hybrid Visual Encoders for Occluded Person Re-Identification}  


\tnotetext[1]{} 

%

\author[1,2] {Zhangjian Ji}
\credit{Conceptualization, Writing – review \& editing, Funding acquisition, Supervision}
\cormark[1]
\ead{jizhangjian@sxu.edu.cn}
\author[1,2] {Shaotong Qiao}
\credit{Methodology, Investigation, Data curation, Visualization, Software, Writing – original draft}
\ead{qiaoshaotong@sxu.edu.cn}
\author {Kai Feng}
\credit{Writing – review \& editing,Formal analysis,Validation}
\author[1,2] {Wei Wei}
\credit{Writing – review \& editing, Funding acquisition, Project administration}

\affiliation[1] {organization={School of Computer \& Information Technology, Shanxi University},
            addressline={Wucheng Rd.92}, 
            city={Taiyuan},
            postcode={030006}, 
            state={Shanxi},
            country={China}}
\affiliation[2] {organization={Key Laboratory of Computational Intelligence and Chinese Information Processing of Ministry of Education},
            addressline={Wucheng Rd.92}, 
            city={Taiyuan},
            postcode={030006}, 
            state={Shanxi},
            country={China}}

\cortext[1]{Corresponding author}



\begin{abstract}
 Occluded person re-identification focuses on  matching partially visible pedestrians across multiple camera views. However, occlusions disrupt body-region cues, thereby complicating cross-view matching. Most person ReID methods built on pretrained vision-language models only focus on enhancing prompt-based feature learning while ignoring the semantic information of occluders. Based on the success of CLIP-ReID, we propose a novel Dual Prompt Learning ReID (DPL-ReID) model for occluded person ReID. It incorporates a Dual Prompt Learning (Dual-PL) strategy, which can utilize textual cues to capture complete pedestrian semantics and keep robustness against occlusion, and a Real-World Occlusion Augmentation (RWOA) method that realistically simulates occlusion scenarios encountered in real word to enrich occluded samples. In addition, we also design a Weighted Gated Feature Fusion (WGFF) method, which in corporates LSNet to capture global information and act as a feature-gating mechanism. This mechanism can effectively guide the CLIP visual encoder toward generating more comprehensive feature representations. Extensive experiments on several benchmark occluded ReID datasets show that our proposed DPL-ReID achieves the state-of-the art performance. The occlusion instance library is available at https://github.com/stone-qiao/DPL-ReID.
\end{abstract}




\maketitle

\section{Introduction}\label{Introduction}



\begin{figure}
    \centering
    \includegraphics[width=0.8\linewidth]{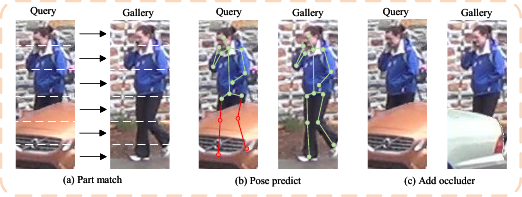}
    \caption{   Current mainstream approaches for handling occluded person ReID: (a) Part-based matching methods, which ignore occlusion information. (b) Methods that introduce human pose cues to predict occluded regions. (c) Methods that add occluders to simulate real-world occlusion.}
    \label{fig:current_mainstream_approaches}
\end{figure}

Person re-identification (ReID) aims to match the same pedestrian across different scenes and camera vipews. In practice scenarios, numerous challenges, such as multi-view variations, pose changes, occlusions, and background noise can significantly degrade matching performance. Current occluded ReID methods mainly fall into three categories.
Part-based matching: focusing on visible body regions \citep{li2021diverse,sun2018beyond,wang2018learning,jia2021matching},as shown in Figure.~\ref{fig:current_mainstream_approaches}(a);
External cues: using pose or other information to predict occluded regions and recover full-body features \citep{gao2020pose,wang2020high,wang2022pose,zheng2021pose}, as shown in Figure.~\ref{fig:current_mainstream_approaches}(b);
Occlusion simulation: augmenting data by random erasing, cropping, or pasting occluders to mimic real-world occlusion, as shown in Figure.~\ref{fig:current_mainstream_approaches}(c).

The first two approaches largely ignore the semantic role of occluders: part-based methods are sensitive to viewpoint and pose changes, while pose-based methods struggle due to low-resolution images in ReID datasets (e.g., Market-1501 \citep{zheng2015person} at 128 × 64) versus the higher input resolution required by pose networks (at least 256 × 192 for HRNet \citep{sun2019deep}, HRFormer \citep{yuan2021hrformer}, ViTPose \citep{xu2022vitpose}). Since most occluded ReID datasets have unoccluded training images and occluded test images, simulating occlusion in training is a more effective way to reduce this domain gap.

\begin{figure}
    \centering
    \includegraphics[width=0.7\linewidth]{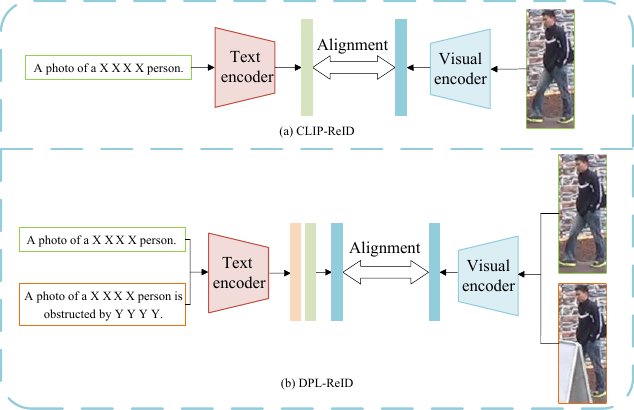}
    \caption{   Two different person re-identification frameworks. (a) The CLIP-ReID framework, which aligns image and text features. (b) Our proposed DP-ReID, equipped with both holistic pedestrian prompts and occluded pedestrian prompts.}
    \label{fig:clipreidanddpreid}
\end{figure}

We revisit the role of occluders in person ReID, focusing on how to better simulate real-world occlusion and how to handle occlusion under CLIP \citep{radford2021learning} vision–language models. For the former, we propose a Real-World Occlusion Augmentation (RWOA) strategy, where each occluder in a realistic occlusion instance library is assigned a fixed and plausible spatial region. For the latter, we introduce a Dual Prompt Learning (Dual-PL) strategy: one prompt learns holistic pedestrian semantics on the original dataset, while the other learns occlusion-aware semantics on the same dataset augmented with occluders. The two are then fused to preserve rich identity semantics while enhancing robustness to occlusion as illustrated in Figure.~\ref{fig:clipreidanddpreid}.
In addition, due to limited receptive fields of visual encoders, we observe that some existing ReID models rely excessively on a single body region for identification. To address this issue, we design a Weighted Gated Feature Fusion (WGFF) method and introduce LSNet \citep{wang2025lsnet}, a lightweigh module that can effectively capture global features. Since we rely on the pre-trained vision–language model CLIP and cannot modify the structure of its visual encoder, we use features from LSNet as a weighted gated fusion with CLIP visual features. This enables the CLIP visual encoder to obtain a broader field of view and attend to more pedestrian regions.

Our main contributions are summarized as follows:
\begin{itemize}
  \itemsep 0pt
  \item We construct an occlusion library consisting of 1,000 occlusion instances and propose a real-world occlusion augmentation method to generate occlusion effects that are not only more realistic but also semantically coherent.
  \item We introduce a dual prompt learning strategy that provides text features that contain rich pedestrian semantics and are resistant to occlusion.
  \item We design a WGFF method that enables the model to extract visual features with a expanding field of view and encompass more comprehensive information.
  \item Extensive experiments on benchmark occluded ReID datasets demonstrate that our approach is superior to state-of-the-art (SOTA) methods.
\end{itemize}





\section{Related Works}\label{Related Works}
\subsection{Pretrained Vision-Language Models}

Pretrained vision–language models (VLMs), especially CLIP, excel at cross-modal alignment by embedding images and text into a shared semantic space. CoOp \citep{zhou2022learning} further enhances CLIP with learnable prompts, inspiring its use in person ReID. CLIP-ReID \citep{li2023clip} uses textual descriptions as high-level cues to learn discriminative pedestrian features. Compared to purely visual models, text prompts provide more stable semantics, improving robustness and generalization.

However, existing prompt learning strategies in CLIP-ReID mainly focus on holistic pedestrian representations and do not explicitly model occlusions. Under severe occlusion, complex backgrounds, or large viewpoint changes, CLIP’s semantic alignment can degrade. Thus, a key challenge is to retain CLIP’s semantic strengths while improving its ability to handle occluded scenarios.

\begin{figure}
    \centering
    \includegraphics[width=0.6\linewidth]{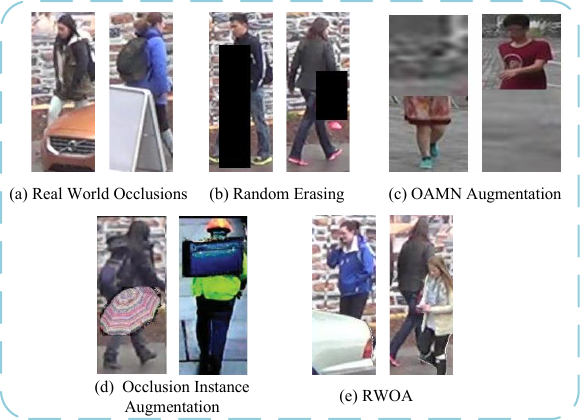}
    \caption{   Examples of real occluded pedestrians and different augmentation strategies. (a) Real-world occlusion scenarios. (b) Simulate occlusion by randomp erasing. (c) OAMN augmentation. (d) OIA augmentation from the FCFormer. (e) Our proposed Real-World Occlusion Augmentation (RWOA).}
    \label{fig:occ}
\end{figure}

\subsection{Occlusion Augmentation Method}

Earlier occluded person re-identification methods mainly focused on part-based feature modeling, matching pedestrians by segmenting the human body or extracting local region features, as well as introducing additional human pose information as prior knowledge. However, the former generally achieves limited performance due to the insufficient visible pedestrian information under occlusion. The latter significantly increases model parameters because of the additional pose estimation network. Moreover, since ReID images usually have relatively low resolution, even after upsampling, severe human keypoint misalignment frequently occurs when they are fed into the pose estimation model.

Recent work has explored occlusion simulation for data augmentation, including simulating occlusion and pasting occluders to increase occluded sample diversity. Random Erasing \citep{zhong2020random}, CutOut \citep{devries2017improved}, and Hide-and-Seek \citep{kumar2017hide} apply non-semantic masks: a rectangle, a square, and multiple small squares, respectively. Random cropping simulates partial missing or viewpoint changes but often removes key identity regions. CutMix \citep{yun2019cutmix} overlays local regions from other pedestrians, disrupting global structure.

Pasting-based methods, such as OAMN \citep{chen2021occlude}, superimpose occluders from a background library onto pedestrian images, but these rarely match real-world patterns. SUREID \citep{song2024deep} employs realistic categories of occluders, such as bicycles, cars and umbrellas, yet represents them as non-instance rectangular patches, which results in visual inconsistency with pedestrians. FED \citep{wang2022feature} uses instance-level occluders but from limited datasets, reducing generalization. FCFormer \citep{wang2024feature} applies Mask R-CNN \citep{he2017mask} to build an occlusion library; however, the placement of occluders often appears unrealistic, even with category-specific placement rules (Figure~\ref{fig:occ}(d)).

Overall, existing methods fail to model occlusion semantics and placement accurately. To address this, we propose a Real-World Occlusion Augmentation (RWOA) method, which uses semantically appropriate occluders at reasonable locations, better simulating realistic occlusions.

\begin{figure}
    \centering
    \includegraphics[width=0.5\linewidth]{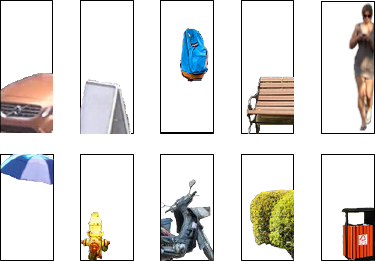}
    \caption{   Some samples from the occlusion instance library.}
    \label{fig:oia}
\end{figure}

\subsection{Prompt Learning Optimization}

Prompt learning is critical for pretrained vision–language models such as CLIP, as it aligns visual features with high-level linguistic semantics to yield transferable representations. RGANet \citep{he2023region} generates region-level semantic prototypes using a CLIP text encoder and aligns them with visual regions to enhance local discrimination. CLIP-ReID adopts CoOp-style learnable prompts and a two-stage training strategy to inject identity information into textual descriptions, thereby improving visual feature learning.

While CLIP-ReID performs well in non-occluded scenarios, it mainly focuses on holistic pedestrian semantics and insufficiently models occlusion. To alleviate this, AG-ReID \citep{zhi2025attribute} filters textual attributes corresponding to the visible regions, and FBA \citep{huang2025background} suppresses background noise via adversarial foreground–background prompts. However, both methods largely ignore the information pertaining to occluders. In fact, occluders can provide explicit cues for occlusion awareness,facilitating more flexible matching between full-body and occluded scenarios.

\begin{figure}
    \centering
    \includegraphics[width=\linewidth]{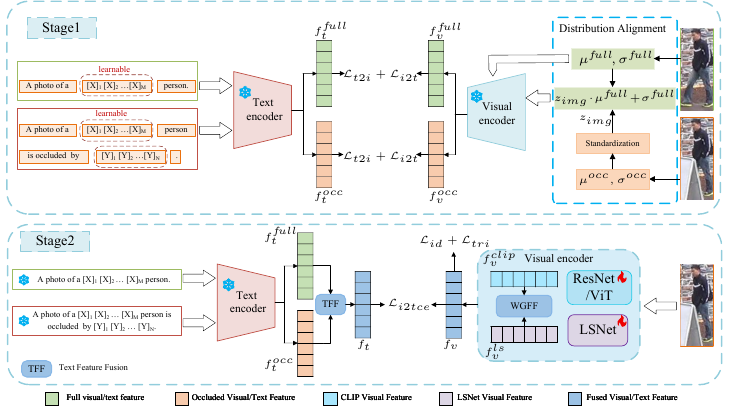}
    \caption{The overall architecture of the dual-prompt learning CLIP hybrid visual encoder (DPL-ReID). DPL-ReID adopts a two-stage training strategy that extends CLIP-ReID with a Dual Prompt Learning module (Dual-PL) and a Weighted Gated Feature Fusion method (WGFF). In the first stage, two sets of image–text pairs are constructed. These two pairs correspond to the full-body semantics and occlusion semantics of pedestrians, respectively. In the second stage, the two types of textual features obtained from the first stage are fused, and a hybrid visual encoder, consisting of the original CLIP visual encoder and an additional LSNet branch being responsible for extracting global features, is introduced.}
    \label{fig:model}
\end{figure}

\begin{figure}
    \centering
    \includegraphics[width=0.6\linewidth]{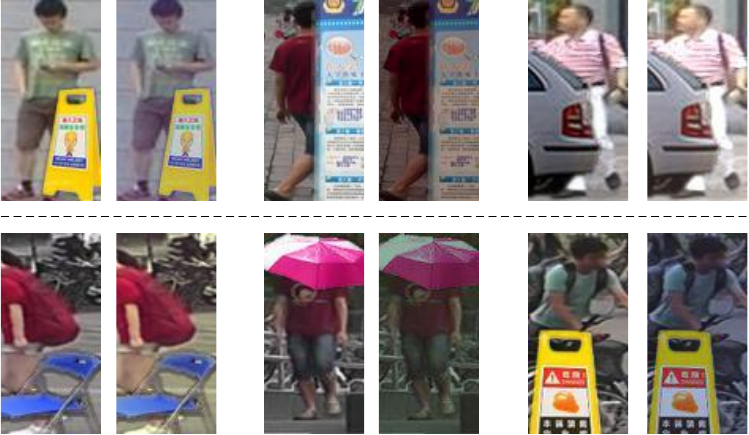}
    \caption{Visualization of the effects of Distribution Alignment on occluded images. The first row adjusts the brightness of the occluder to match the surrounding environment. The second row reduces the contrast of the occluder relative to the background.}
    \label{fig:alignment}
\end{figure}

\section{Method}
In this section, we provide a detailed introduction to our dual-prompt learning hybrid encoder framework based on CLIP-ReID (DPL-ReID). We first adopt a new occlusion augmentation strategy to simulate more realistic real-world occlusions. Then, we propose a dual prompt learning mechanism that integrates both full-body textual descriptions and occlusion-aware textual descriptions, enabling the model to inherit rich pedestrian semantics while maintaining robustness under occlusion. Additionally, to overcome the limitations of CLIP’s image encoder, namely its inadequate feature extraction ability and limited receptive field, we design a hybrid image encoder strategy. Specifically, we integrate LSNet, which can extract robust global features, and use its global representations as a gating signal to refine the CLIP image features.

\subsection{Real-World Occlusion Augmentation}

Although most existing occlusion augmentation strategies outperform random erasing or random cropping methods that lack occluder semantics, they still fail to simulate realistic real-world occlusion due to inappropriate occluder selection and unnatural placement. To tackle this issue, we build a realistic occlusion instance library containing semantically meaningful occluders and devise a more rational occlusion processing strategy.

We meticulously select occluders, as this step is critical for achieving realistic augmentation. Occluded person e-identification mainly occurs in outdoor scenes, such as shopping streets and campuses, rather than in indoor environments. Accordingly, outdoor occluders can be classified into two categories: strong occluders that cover over 40\% of the pedestrian region (e.g., cars, trees, walls), and weak occluders with smaller coverage (e.g., road barriers, fire hydrants, carried objects such as backpacks, handbags, umbrellas, signboards, and mild inter-person occlusion). In contrast, some methods use indoor objects (e.g., tables or chairs) to paste onto pedestrian images, which will introduce semantic conflicts. We therefore extract occluders from person ReID datasets (e.g., Market-1501 and DukeMTMC-reID \citep{zheng2017unlabeled} ), object classification datasets (e.g.,COCO2017 \citep{lin2014microsoft}  and BDD100K \citep{yu2020bdd100k} ) and images collected from the web, to build an occlusion instance library containing 1,000 images, We manually refined and extracted the occluders to produce smoother and more realistic occlusion instances, instead of directly adopting the coarse segmentation results generated by Mask R-CNN, which usually contain redundant background regions and incomplete object structures.We manually refined and extracted the occluders to produce smoother and more realistic occlusion instances, instead of directly adopting the coarse segmentation results generated by Mask R-CNN, which usually contain redundant background regions and incomplete object structures ( as shown in Figure~\ref{fig:oia}).

We further observe that pedestrian images generally contain a complete human body, the pedestrian occupies a relatively consistent proportion of the entire image, and occluders often appear at fixed spatial regions (e.g., cars usually occlude the lower body and are only partially visible). Random placement may result in illogical semantic occlusions, such as a partial car being pasted above the waistline of a pedestrian. To resolve this, we preserve the original size of each occluder image, setting non-occluded regions to transparent to ensure correct spatial semantics. The occlusion instance is then resized to match the pedestrian image, its boundaries are feathered to reduce sharp edges, and finally overlaid. Some examples of the occlusion instance image are shown in Figure ~\ref{fig:oia}.

\begin{figure}
    \centering
    \includegraphics[width=\linewidth]{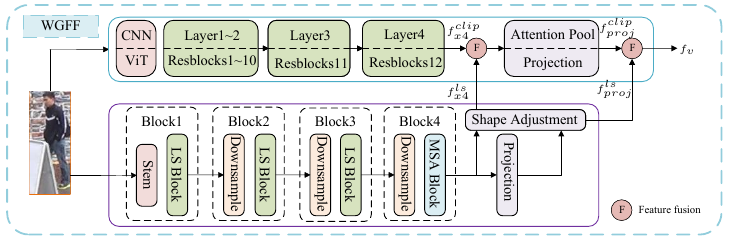}
    \caption{Weighted gated feature fusion. Different fusion strategies are applied for CLIP visual encoders with different backbones.}
    \label{fig:wgff}
\end{figure}

\subsection{Dual Prompt Learning Strategy}

Current vision–language ReID models, such as CLIP-ReID, learn only holistic textual prompts for pedestrians and ignore the role of occluders in occluded scenarios.

Let $i \in\{1\dots B\}$ index a mini-batch image. In Stage 1, the image $img_{i}$ is fed into the frozen CLIP image encoder $\mathcal{I}(\cdot)$ to obtain visual features $f^{i}_{v}$. The text descriptions $text_{y_{i}}$ (since each identity share the same textual description, $y_{i}$ denotes the pedestrian ID corresponding to the $i$th image) that input the frozen CLIP text encoder $\mathcal{T}(\cdot)$ are defined as  “A photo of a $\text{[X]}_{\text{1}}\ \text{[X]}_{\text{2}}\ \cdots\ \text{[X]}_{\text{M}}$ person." where each $[\text{X}]_{\text{m}} \;(\text{m} \in 1,\dots\text{M})$ is a learnable token initialized by the pedestrian ID. A bidirectional image–text contrastive loss optimizes the tokens to capture ID-specific semantics. In Stage 2, these textual features guide the image encoder to better extract pedestrian features.

To explicitly incorporate occluder information, we propose a Dual Prompt Learning (Dual-PL) framework, which learns two types of prompts: one capturing holistic pedestrian semantics and the other modeling occlusion semantics, through Full-body and Occlusion templates.

\[
\begin{aligned}
\text{Template}^{full} &: 
\text{“A photo of a } [\text{X}]_1 [\text{X}]_2 \cdots [\text{X}]_M \text{ person.”} \\
\\
\text{Template}^{occ} &: 
\text{“A photo of a } [\text{X}]_1 [\text{X}]_2 \cdots [\text{X}]_M \text{ person is occluded by } [\text{Y}]_1 [\text{Y}]_2 \cdots [\text{Y}]_N \text{.”}
\end{aligned}
\]

Both $\text{[X]}_\text{m}$ and $\text{[Y]}_\text{n}$ are learnable tokens, where $m \in\{1\dots \text{M}\}$ and $n \in\{1\dots \text{N}\}$.
We denote the corresponding textual inputs as $text^{full}_{y_{i}}$ and $text^{occ}_{y_{i}}$.

Training the full-body semantics, The clean image $img^{full}_{i}$ and $text^{full}_{y_{i}}$ are fed into the frozen CLIP encoders to obtain the full-body visual feature $f^{i,full}_{v}$ and text feature $f^{y_{i},full}_{t}$.
The bidirectional contrastive losses $\mathcal{L}_{t2i}$ and $\mathcal{L}_{i2t}$ are applied to optimize the tokens $\text{[X]}_\text{m}$.The $\mathcal{L}_{t2i}$ and $\mathcal{L}_{i2t}$ losses are computed as follows:

\begin{equation}
\mathcal{L}_{i2t}(i)
= - \log \frac{\exp\left(s\!\left(f^{i}_{v}, f^{y_i}_{t}\right)\right)}
{\sum_{a=1}^{B} \exp\left(s\!\left(f^{i}_{v}, f^{a}_{t}\right)\right)} ,
\end{equation}

\begin{equation}
\mathcal{L}_{t2i}(y_i)
= - \frac{1}{\left|\mathcal{P}(y_i)\right|}
\sum_{p \in \mathcal{P}(y_i)}
\log
\frac{\exp\!\left(s\!\left(f^{p}_{v}, f^{y_i}_{t}\right)\right)}
{\sum_{a=1}^{B} \exp\!\left(s\!\left(f^{a}_{v}, f^{y_i}_{t}\right)\right)}
\end{equation}

$\mathcal{P}(y_i) = \{\, p \in 1\dots B : y_p = y_i \,\}$ denotes the index set of all positive samples in the mini-batch.

We further consider the appearance consistency between occluders and the original pedestrian images. Since occluding instances are collected from diverse sources, they exhibit large variations in brightness, saturation, and sharpness, and often differ significantly from the target pedestrian images. Directly pasting occluders onto pedestrian regions can easily introduce obvious style inconsistencies, thereby causing additional appearance distribution shift and adversely affecting identity feature learning.

This issue becomes even more pronounced in CLIP-based person ReID frameworks. As the CLIP visual encoder is pre-trained on large-scale natural image–text pairs, its feature representations are highly sensitive to global visual distributions. When occluders with distinct style discrepancies are directly overlaid onto pedestrian images, they may disrupt the original natural image statistics, leading to abnormal local semantic responses and interfering with identity-discriminative feature learning.

To mitigate this problem, we draw inspiration from AdaIN \citep{Huang2017ArbitraryST} and perform lightweight alignment on the synthesized occluded images. By adjusting their mean and variance, we enforce consistency with the appearance distribution of the original pedestrian images, thereby improving the realism and scene consistency of occlusion generation. As shown in Figure ~\ref{fig:alignment}, after alignment, the occluded regions can be more naturally integrated into the original images. Moreover, the overall brightness variations can simulate real-world illumination changes and further enhance data diversity.
We first standardize:
\begin{equation}
z_{img} = \frac{img^{occ}_{i} - \mu^{occ}}{\sigma^{occ}}
\end{equation}
and then match it to the clean-image distribution: 
\begin{equation}
img^{occ}_{i} = z_{img} \cdot \sigma^{full} + \mu^{full}
\end{equation}

The adjusted occluded image $img^{occ}_{i}$ and the occlusion textual prompt $text^{occ}_{y_{i}}$ are passed into CLIP to obtain $f^{i,occ}_{v}$ and $f^{y_i,occ}_{t}$.
The bidirectional contrastive losses optimize both $\text{[X]}_\text{m}$ and $\text{[Y]}_\text{n}$.
The two prompt-learning branches are trained independently without parameter sharing. The final loss in the first stage is defined as follows:

\begin{equation}
\mathcal{L}_{stage1}
= \mathcal{L}_{i2t}^{full}
+ \mathcal{L}_{t2i}^{full}
+ \mathcal{L}_{i2t}^{occ}
+ \mathcal{L}_{t2i}^{occ}
\end{equation}

In Stage 2, we learn a weighted fusion of the full-body text feature $f^{y_i,full}_{t}$ and the occlusion text feature $f^{y_i,occ}_{t}$
\begin{equation}
f^{y_i}_{t} = \alpha f^{y_i,full}_{t} + (1-\alpha) f^{y_i,occ}_{t}
\end{equation}

The $\alpha$ is a learnable parameter and is used to fuse text feature $f^{y_i}_{t}$ that contains rich holistic semantics and is robust to occlusion. Then, it is also used to guide the image encoder in extracting more discriminative visual features.

\subsection{Weighted Gated Feature Fusion}

Most current ReID models often rely on a specific local region to serve as the unique identifier for pedestrians. As shown in the left part of Figure~\ref{fig:cam}, CLIP-ReID may focus on partial cues (e.g., shoulders), which can lead to misidentification if that region is occluded. This phenomenon arises because the visual encoder has a limited receptive field, and modifying CLIP’s pretrained weights or architecture would break its image–text alignment, thereby preventing the acquisition of reliable textual features in Stage~1.

To address this, we propose a Weighted Gated Feature Fusion (WGFF) mechanism ( as shown in Figure ~\ref{fig:wgff}). 
To effectively capture global pedestrian features, in WGFF mechanism, we introduce a visual encoder (LSNet) with a large receptive field, which uses large-kernel convolutions for broad context “See Large” and small-kernel, attention-guided convolutions for fine details “Focus Small”. These global features are further fused with the features derived from the CLIP visual encoder to provide richer and more robust visual representations for person ReID. In addition, considering the different output representations of the CLIP visual encoder, which produces spatial feature maps (H×W×C) in CNN-based architectures and token-level linear embeddings in ViT-based architectures, we design different fusion strategies for each type.

For CLIP models that use the ResNet-50 \citep{he2016deep} visual encoder, we adopt the following gated feature fusion mechanism:
\begin{equation}
f_{x4} = f^{clip}_{x4} \cdot \sigma\!\left(f^{ls}_{x4}\right) 
\end{equation}
\begin{equation}
f_{v} = f^{clip}_{proj} \cdot \sigma\!\left(f^{ls}_{proj}\right) 
\end{equation}

Here, $f^{clip}_{x4}$ and $f^{ls}_{x4}$ are features extracted from CLIP’s Layer 4 and LSNet’s Block 4, respectively, and $f^{clip}_{proj}$ and $f^{ls}_{proj}$ are their final pooled projection features.  $\sigma(\cdot)$is a sigmoid activation function. Before fusion, $f^{ls}_{x4}$ is reshaped to match the dimension of $f^{clip}_{x4}$. LSNet's global features guide the CLIP features to capture more comprehensive person-related information. The same applies to $f^{clip}_{proj}$ and $f^{ls}_{proj}$.

For CLIP models with a ViT-B \citep{dosovitskiy2020image} visual encoder, since Transformer architectures inherently possess a relatively large receptive field, we adopt a learnable weighted fusion strategy:
\begin{equation}
f_{x4} = \beta\, f^{clip}_{x4} + (1-\beta)\, f^{ls}_{x4}
\end{equation}
\begin{equation}
f_{v} = \beta\, f^{clip}_{proj} + (1-\beta)\, f^{ls}_{proj}
\end{equation}
where $\beta \in [0,1]$ is a learnable weight parameter. In this case, 
$f^{clip}_{x4}$ and $f^{clip}_{proj}$ correspond to the features 
from ViT\text{-}B's 12th residual block and its projection layer.

In the second stage, we follow CLIP-ReID and adopt the ID loss, triplet loss, and contrastive loss. The corresponding formulas are given as follows:
\begin{equation}
\mathcal{L}_{id} = - \sum_{k=1}^{N} q_k \log(p_k),
\end{equation}

\begin{equation}
\mathcal{L}_{tri} = \max(d_p - d_n + \alpha, 0),
\end{equation}

\begin{equation}
\mathcal{L}_{i2tce}(i)
= - \sum_{k=1}^{N} q_k
\log
\frac{
\exp\left(s\left(f^{i}_{v}, f^{y_k}_{t}\right)\right)
}{
\sum_{y_a=1}^{N}
\exp\left(s\left(f^{i}_{v}, f^{y_a}_{t}\right)\right)
}.
\end{equation}

\begin{equation}
\mathcal{L}_{stage2}
= \mathcal{L}_{id}
+ \mathcal{L}_{tri}
+ \mathcal{L}_{i2tce}
\end{equation}

Here, $q_k = (1 - \epsilon)\delta_{k,y_i} + \epsilon / N$ denotes the
value in the target distribution with label smoothing, and $p_k$
represents the predicted ID logits of class $k$. The terms $d_p$ and
$d_n$ denote the distances of positive and negative pairs,
respectively, while $\alpha$ is the margin of the triplet loss.
Label smoothing is applied to $q_k$ in
$\mathcal{L}_{i2tce}$.

\section{Experiments}
\subsection{Datasets and Evaluation Metrics}

To validate the effectiveness of our method, we evaluate it on four occluded ReID datasets: Occluded Duke \citep{miao2019pose}, Occluded-REID \citep{zhuo2018occluded}, P-DukeMTMC \citep{zhuo2018occluded}, and Occluded-Market \citep{han2023spatial}. The details of these datasets are summarized in Table \ref{tab:dataset_statistics}.
\begin{table}
\caption{Statistics of occluded person re-identification datasets. ``*'' indicates that Occluded-REID does not provide a dedicated training set, and the training set of Market-1501 is adopted for training.}
\centering
\begin{tabular}{l c cc cc cc}
\toprule
\multirow{2}{*}{Dataset} & \multirow{2}{*}{Cam} 
& \multicolumn{2}{c}{Training dataset} 
& \multicolumn{2}{c}{Query} 
& \multicolumn{2}{c}{Gallery} \\
\cmidrule(lr){3-4} \cmidrule(lr){5-6} \cmidrule(lr){7-8}
& & ID & Image & ID & Image & ID & Image \\
\midrule
Occluded-Duke   & 8 & 702 & 15618 & 702 & 2210 & 702 & 17661 \\
Occluded-REID*  & 6 & \underline{751} & \underline{12936} & 200 & 1000 & 200 & 1000 \\
P-DukeMTMC      & - & 665 & 12927 & 634 & 2163 & 634 & 9053 \\
Occluded-Market & - & 780 & 9287  & 533 & 2343 & 751 & 15913 \\
\bottomrule
\end{tabular}
\label{tab:dataset_statistics}
\end{table}

For the fairness comparison, we follow conventions in the ReID community and adopt the Cumulative Matching Characteristic (CMC) curve and mean Average Precision (mAP) to evaluate different ReID models. Specifically,
we adopt the first element of the CMC curve, i.e., Rank-1, which indicates the hit rate of the first match.

\subsection{Implementation Details}

Our encoder follows CLIP-ReID, with ResNet50 and ViT-B/16 as visual backbones. All input images are resized to $256 \times 128$. We adopt a two-stage training strategy with the Adam optimizer and run on a single NVIDIA A100 GPU.

In the first stage, the learning rate is $3.5 \times 10^{-4}$ with cosine decay, batch size is 64, no traditional data augmentation is applied, and LSNet is not used. Training lasts 120 epochs for CNNs and 150 for ViTs. In the second stage, each mini-batch contains $B = P \times K$ images with $P=16$ identities and $K=4$ samples per identity. Images are augmented with random flipping, padding, cropping, and erasing to reduce overfitting. For CNN, the learning rate warms up from $3.5 \times 10^{-6}$ to $3.5 \times 10^{-4}$ over 10 epochs and decays by 0.1 at epochs 40 and 70; the model converges around 200 epochs. For ViT, it warms up from $5 \times 10^{-7}$ to $5 \times 10^{-6}$ over 10 epochs, decays by 0.1 at the 30th and 50th epochs, and trains for 80 epochs. LSNet is integrated in this stage. 

\subsection{Comparison with State-of-the-Art Methods}
\begin{table}
\centering 
\caption{Performance comparison with state-of-the-art methods on occluded ReID benchmarks. “*” indicates that the model is initialized with ImageNet pretraining. 
“${\ddagger}$” indicates that the model is initialized with LUPerson pretraining.} 

\resizebox{\linewidth}{!}{
\begin{tabular}{c l l c c c c c c c c} 
\toprule 
\multirow{2}{*}{Backbone} & \multirow{2}{*}{Methods} & \multirow{2}{*}{Reference} & \multicolumn{2}{c}{Occluded-Duke} & \multicolumn{2}{c}{Occluded-REID} & \multicolumn{2}{c}{P-DukeMTMC} & \multicolumn{2}{c}{Occluded-Market} \\ 
& & & mAP & R1 & mAP & R1 & mAP & R1 & mAP & R1 \\ 
\midrule

\multirow{6}{*}{CNN} & PCB \citep{sun2018beyond} & ECCV 18 & 33.7 & 42.6 & 38.9 & 41.3 & 63.9 & 79.4 & - & - \\ 
& HOReID  \citep{wang2020high}  & CVPR 20 & 43.8 & 55.1 & \textbf{70.2} & \textbf{80.3} & - & - & 49.3 & 64.9 \\ 
& PVPM \citep{gao2020pose} & CVPR 20 & 37.7 & 47.0 & 61.2 & 70.4 & 69.9 & 85.1 & 49.4 & 66.8 \\ 
& OAMN \citep{chen2021occlude} & ICCV 21 & 46.1 & 62.6 & - & - & - & - & - & - \\ 
\cmidrule[0.4pt]{2-11} 
& CLIP-ReID \citep{li2023clip} & AAAI 23 & 53.5 & 61.0 & 52.9 & 58.6 & \textbf{81.3} & 92.0 & \textbf{65.6} & 79.8 \\ 
& DPL-ReID & ours & \textbf{57.2} & \textbf{70.6} & 62.4 & 70.8 & 81.2 & \textbf{92.4} & 65.1 & \textbf{80.8} \\ 
\midrule 
\multirow{14}{*}{ViT} & TransReID \citep{he2021transreid} & ICCV 21 & 55.7 & 64.2 & 67.3 & 70.2 & - & - & 69.7 & 80.2 \\ 
& PAT \citep{li2021diverse} & CVPR 21 & 53.6 & 64.5 & 72.1 & 81.6 & - & - & - & - \\ 
& FED \citep{wang2022feature} & CVPR 22 & 56.4 & 68.1 & 79.3 & 86.3 & - & - & 53.3 & 66.7 \\ 
& RGANet \citep{he2023region} & TIFS 23 & 62.4 & 71.6 & 80.0 & 86.4 & - & - & - & - \\ 
& OSRTrans \citep{10.1016/j.neucom.2024.127442} & NC 24 &  61.5 & 72.9 & - & - & - & - & 67.5 & 82.0 \\ 
& FCFormer \citep{wang2024feature} & TMM 24 & 63.1 & 73.0 & 86.2 & 84.9 & 82.5 & 92.4 & - & - \\ 
& SSSC-TransReID \citep{Ji2025Exploring} & MMS 25 & 61.0 & 69.2 & - & - & - & - & 71.7 & 83.3 \\ 
& AG-ReID \citep{zhi2025attribute} & ECAI 25 & 63.2 & 70.4 & 87.6 & 90.1 & 84.5 & 91.8 & - & - \\ 
& THCB-Net* \citep{wang2025looking} & \multirow{2}{*}{TIFS 25} & 62.6 & 72.3 & 84.5 & 87.3 & - & - & - & - \\ 
& THCB-Net${\ddagger}$ & & \textbf{70.7} & \textbf{80.0} & 84.8 & 88.9 & - & - & - & - \\ 
& FBA \citep{huang2025background} & arXiv 25 & 60.5 & 69.5 & 84.0 & 85.4 & - & - & - & - \\ 
\cmidrule[0.4pt]{2-11} 
& CLIP-ReID \citep{li2023clip} & AAAI 23 & 60.3 & 67.2 & 88.7 & 91.8 & 83.4 & 92.2 & 71.4 & 83.8 \\ 
& DPL-ReID & ours & 67.2 & 76.3 & \textbf{91.3} & \textbf{94.9} & \textbf{85.9} & \textbf{94.5} & \textbf{74.4} & \textbf{86.6} \\ 
\bottomrule 
\end{tabular} 
}
\label{tab:occ_result} 
\end{table}
Table~\ref{tab:occ_result} presents results on four occluded datasets, comparing both CNN- and ViT-based ReID methods. Our DPL-ReID consistently outperforms existing state-of-the-art approaches.

For CNN backbones, DPL-ReID significantly outperforms CLIP-ReID, achieving 57.2\% mAP and 70.6\% Rank-1 on Occluded-Duke, and 62.4\% mAP and 70.8\% Rank-1 on Occluded-REID. For P-DukeMTMC and Occluded-Market, their training sets already contain a large proportion of occluded images, so we reduce the occlusion augmentation ratio, resulting in less pronounced performance improvements.

For ViT backbones, DPL-ReID achieves state-of-the-art performance on most datasets. On Occluded-Duke, it attains 67.2\% mAP and 76.3\% Rank-1, second only to THCB-Net (based on Swin-B \citep{liu2021swin}) pre-trained on the LUPerson \citep{fu2021unsupervised} dataset that consists of 4M pedestrian images of over 200 identities, but outperforming THCB-Net pre-trained on ImageNet \citep{russakovsky2015imagenet}. On Occluded-REID, DPL-ReID reaches 91.3\% mAP and 76.3\% Rank-1, surpassing previous methods. Consistent gains are also observed on P-DukeMTMC (85.9\% mAP, 94.5\% Rank-1) and Occluded-Market (74.4\% mAP, 86.6\% Rank-1).

\subsection{Ablation Studies and Analysis}

\textbf{Effectiveness of the proposed modules.}
We perform ablation studies on Occluded-Duke and Occluded-REID using CLIP-ReID with a ViT-B/16 backbone, stride size $12 \times 12$, and both SIE and OLP as the baseline.

As shown in Table~\ref{tab:ablation}, adding the weighted gated feature fusion (WGFF) module improves performance by 0.9\% mAP and 0.7\% Rank-1 on Occluded-Duke, and 0.4\% mAP on Occluded-REID, indicating that LSNet guides the CLIP visual encoder to capture broader and richer pedestrian features. When only add our real-world occlusion augmentation (RWOA) into the baseline, the mAP and Rank-1 scores increase by 5.2\% and 7.9\% on Occluded-Duke, and improve by 1.6\% and 1.7\% on Occluded-REID. The reason is that increasing occluded samples reduces the domain gap between training and testing sets. Continuing to integrate dual prompt earning (Dual-PL) into the baseline, the mAP and Rank-1 scores further raise by 1.2\% and 1.1\% on Occluded-Duke, and also yield gains of 0.9\% and 1.3\% on Occluded-REID, as the Dual-PL fuses textual features from two contextual scenarios. Finally, by integrating all three modules, the full model achieves the optimal performance on both Occluded-Duke and Occluded-REID.

\begin{table}
\centering
\caption{Ablation study on component effectiveness}
\begin{tabular}{c c c c c c c}
\toprule
\multicolumn{3}{c}{Components} & \multicolumn{2}{c}{Occluded-Duke} & \multicolumn{2}{c}{Occluded-REID} \\
\cmidrule(lr){1-7}
RWOA & Dual-PL & WGFF & mAP & R1 & mAP & R1 \\
\midrule
- & - & - & 60.3 & 67.2 & 88.7 & 91.8 \\
- & - & $\surd$ & 61.2 & 67.9 & 89.1 & 91.6 \\
$\surd$ & - & - & 65.7 & 75.1 & 90.3 & 93.5 \\ 
$\surd$ & $\surd$ & - & 66.9 & 76.2 & 91.2 & 94.8 \\
$\surd$ & $\surd$ & $\surd$ & \textbf{67.2} & \textbf{76.3} & \textbf{91.3} & \textbf{94.9} \\
\bottomrule
\end{tabular}
\label{tab:ablation}
\end{table}

\begin{figure}
    \centering
    \includegraphics[width=0.6\linewidth]{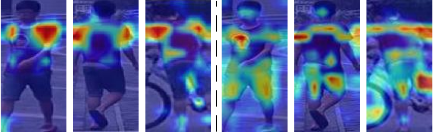}
    \caption{   Grad-CAM visualization. Left: ResNet50-based CLIP-ReID; Right: CLIP-ReID with LSNet.}
    \label{fig:cam}
\end{figure}

Simulated occlusion is not suitable for holistic person ReID, as it only enlarges the domain gap between the training and testing sets. We only introduce the Weighted Gated Feature Fusion (WGFF) module into CLIP-ReID. The corresponding results are reported in Table~\ref{tab:full_result}. After adding the proposed WGFF enhancement module, CLIP-ReID for both CNN-based and ViT-based frameworks achieves a certain improvements on Market-1501 and DukeMTMC. Although the performance gain brought by WGFF is relatively smaller compared with other modules, it can effectively alleviate the issue in the original CLIP-ReID where the model tends to regard a local pedestrian region as the sole identity cue, especially in CNN-based frameworks, as illustrated in Figure~\ref{fig:cam}.
\begin{table}
\centering
\caption{Performance of Weighted Gated Feature Fusion (WGFF) on Holistic ReID datasets}

\begin{tabular}{l c c c c}
\toprule
\multirow{2}{*}{Methods} & \multicolumn{2}{c}{Market-1501} & \multicolumn{2}{c}{DukeMTMC} \\
& mAP & R1 & mAP & R1 \\
\midrule
\textit{cnn} & & & & \\ 
PCB \citep{sun2018beyond} & 77.4 & 92.3 & 66.1 & 81.8 \\
MVPM \citep{Sun2019MVPMA} & 80.5 & 91.4 & 70.0 & 83.4 \\
AANet \citep{Tay2019AANetAA} & 82.5 & 93.9 & 72.6 & 86.4 \\
HOReID \citep{wang2020high} & 84.9 & 94.2 & 75.6 & 86.9 \\
\cmidrule[0.4pt]{1-5}
CLIP-ReID \citep{li2023clip} & 89.8 & 95.7 & 80.7 & 90.0 \\
+ WFGG (ours) & \textbf{90.5} & \textbf{96.1} & \textbf{81.5} & \textbf{90.4} \\
\midrule
\textit{vit} & & & & \\ 
TransReID \citep{he2021transreid} & 88.2 & 95.0 & 80.6 & 89.6 \\
PAT \citep{li2021diverse} & 88.0 & 95.4 & 78.2 & 88.8 \\
FED \citep{wang2022feature} & 86.3 & 95.0 & 78.0 & 89.4 \\
PFD \citep{wang2022pose} & 89.6 & 95.5 & 82.2 & 90.6 \\
RGANet \citep{he2023region} & 89.7 & 95.5 & - & - \\
AAFormer \citep{Zhu2021AAformerAT} & 88.0 & 95.4 & 80.9 & 90.6 \\
FCForme \citep{wang2024feature} & 86.8 & 95.0 & 78.8 & 89.7 \\
AG-ReID \citep{zhi2025attribute} & \textbf{90.8} & \textbf{95.8} & 83.3 & 91.0 \\
FBA \citep{huang2025background} & - & - & 81.7 & 91.5 \\
\cmidrule[0.4pt]{1-5}
CLIP-ReID \citep{li2023clip} & 90.5 & 95.4 & 82.5 & 90.0 \\
+ WFGG (ours) & 90.6 & 95.7 & \textbf{83.3} & \textbf{91.1} \\
\bottomrule
\end{tabular}
\label{tab:full_result}
\end{table}

In person ReID, the model should focus on as much pedestrian information as possible, rather than relying on only a specific local region to “trick” the model into believing that the identification task has been completed. WGFF helps address this issue and also brings consistent improvements on both occluded and holistic person ReID datasets. Therefore, we believe that WGFF is meaningful and valuable.

\begin{table}
\centering
\caption{Comparison of different augmentation methods on Occluded-Duke}
\begin{tabular}{l c c c c}
\toprule
Methods & mAP & R1 & R5 & R10 \\
\midrule
\textit{w/o traditional augmentation} & & & & \\
Random Erasing & 60.3 & 66.2 & 80.5 & 85.1 \\
Random Cropping & 58.0 & 66.6 & 80.4 & 84.5 \\
Hide-and-Seek & 58.8 & 65.3 & 80.1 & 84.5 \\
CutMix & 48.4 & 57.5 & 72.9 & 79.1 \\
OIA & 64.9 & 74.8 & 86.7 & 90.0 \\
RWOA(ours) & \textbf{65.7} & \textbf{75.9} & \textbf{87.6} & \textbf{91.0} \\
\midrule
\textit{w traditional augmentation} & & & & \\
Hide-and-Seek & 60.5 & 66.9 & 81.0 & 85.7 \\
CutMix & 52.2 & 59.5 & 76.1 & 82.0 \\
OIA & 66.5 & 74.5 & 87.2 & 90.5 \\
RWOA(ours) & \textbf{66.9} & \textbf{76.2} & \textbf{87.9} & \textbf{91.7} \\
\bottomrule
\end{tabular}
\label{tab:augmentation}
\end{table}

\textbf{Impact of different occlusion augmentation strategies.} To evaluate different occlusion augmentation strategies, Table~\ref{tab:augmentation} gives the comparison results when different augmentations are used in our proposed Dual-PL module to generate the corresponding occlusion image for the occlusion prompt text template, where each augmentation method adds or simulates occlusion at a 50\% ratio. In Table~\ref{tab:augmentation}, we also discuss the cases whether using the traditional augmentation methods (e.g.,random erasing, random cropping,etc) in Stage 2 or not.

Hide-and-Seek introduces fragmented occlusion with multiple small patches, while real-world occlusion usually involves a single large region. CutMix replaces regions with patches from other pedestrians, disrupting global structure and semantic consistency. Both two augmentations make it difficult to obtain semantically complete pedestrian texts and are unsuitable for text-enhanced ReID. OIA, proposed in FCFormer, adopts a semi-random occluder placement strategy. Using the same self-constructed occlusion instance library, we compare OIA’s placement strategy with our fixed-position strategy. Results show that our approach consistently performs better, demonstrating that more reasonable occluder placement preserves pedestrian semantics, reduces unnatural composites, and improves text–visual alignment in cross-modal ReID.

\begin{figure}
    \centering
    \includegraphics[width=0.5\linewidth]{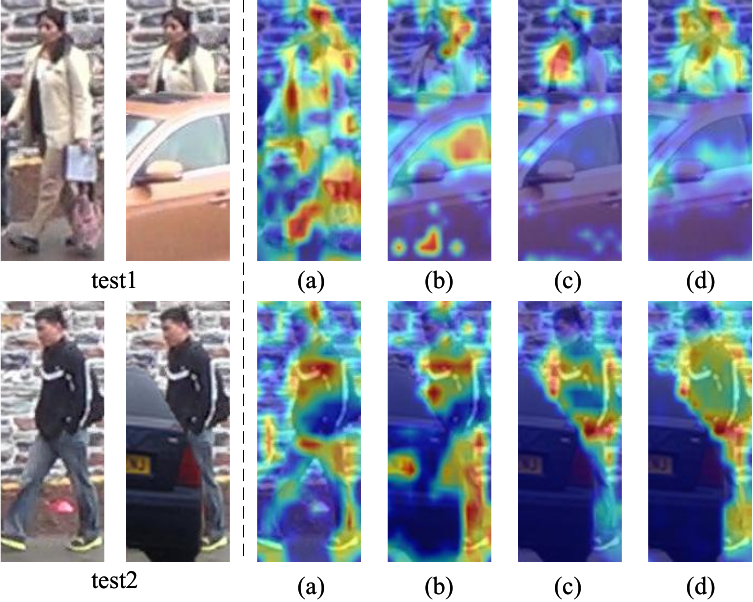}
    \caption{Visualization results of different prompt templates. (a) Template$^{full}$ on full-body pedestrian images. (b) Template$^{full}$ on occluded pedestrian images. (c) Template$^{occ}$ on occluded pedestrian images. (d) Fusion of Template$^{full}$ and Template$^{occ}$ on occluded pedestrian images.}
    \label{fig:template_Visualization}
\end{figure}

\textbf{Impact of different prompt templates.} In order to evaluate the effectiveness of combining two prompt templates in Dual-PL, We compare the combination of the two prompt templates with using either template alone in Table~\ref{tab:prompt_templates}, and further visualize their attention maps using Grad-CAM  \citep{selvaraju2017grad}, as shown in Figure.~\ref{fig:template_Visualization}. The results in Table~\ref{tab:prompt_templates} and Figure.~\ref{fig:template_Visualization} show that Template$^{full}$ can capture richer pedestrian appearance features on full-body pedestrian images (see Figure.~\ref{fig:template_Visualization}(a)). However, once occluders are introduced, the model additionally focuses on some non-pedestrian regions. For example, in the lightly occluded sample test2, where only one leg is occluded, Template$^{full}$ attends to irrelevant regions such as the vehicle license plate and headlights in Figure.~\ref{fig:template_Visualization}(b). In the heavily occluded sample test1, where more than half of the pedestrian body is occluded, pedestrian localization is severely affected.

In contrast, Template$^{occ}$ employs specially designed prompt templates to utilize occluders for assisting pedestrian region localization, thereby achieving higher Rank-1 accuracy. However, since the appearance features are learned from occluded pedestrian datasets, and the occluded regions vary across different samples of the same identity, some pedestrian appearance information may be missing, as illustrated in Figure.~\ref{fig:template_Visualization}(c).

By learning Template$^{full}$ on full-body pedestrian images and Template$^{occ}$ on occluded pedestrian images separately, and then fusing the two textual features, the proposed method can not only preserve richer pedestrian appearance representations but also leverage occluders to facilitate pedestrian localization (see Figure.~\ref{fig:template_Visualization}(d)). As a result, the robustness of the model under occlusion scenarios is significantly improved. 

\begin{table}
\centering
\caption{Comparison of prompt templates on Occluded-Duke}
\begin{tabular}{l c c c c}
\toprule
Prompt template & mAP & R1 & R5 & R10 \\
\midrule
Template$^{full}$ & 65.7 & 75.1 & 87.7 & 91.0 \\
Template$^{occ}$ & 65.8 & 75.9 & 87.7 & 91.1 \\
Both & \textbf{66.9} & \textbf{76.2} & \textbf{87.9} & \textbf{91.7} \\
\bottomrule
\end{tabular}
\label{tab:prompt_templates}
\end{table}

\textbf{Impact of different text feature fusion methods.} In Table~\ref{tab:text_feature_fusion}, we compare three different text feature fusion strategies: linear weighted fusion, a three-layer MLP, and cross-attention fusion. Results show that linear weighted fusion performs best.

The reason is that, during Stage~1, the two text prompts have already been sufficiently trained on full and occluded pedestrian images through contrastive learning, and the textual features have been aligned with their corresponding image features in a shared feature space. Based on this, the objective of Stage~2 is not to remodel the textual semantic space, but rather to introduce robustness under occlusion while preserving the original alignment structure as much as possible. MLP fusion introduces nonlinear transformations that disrupt the text–image geometry, and cross-attention partially preserves features but still modifies relative relationships. However, learning a single scalar weight for linear fusion effectively preserves the textual feature structure while integrating complementary information, thereby achieving optimal performance.

\begin{table}
\centering
\caption{Comparison of three different text feature fusion methods on Occluded-Duke}
\begin{tabular}{l c c c c}
\toprule
Text fusion method & mAP & R1 & R5 & R10 \\
\midrule
Weighted fusion & \textbf{66.9} & \textbf{76.2} & \textbf{87.9} & \textbf{91.7} \\
MLP & 64.3 & 74.0 & 86.4 & 89.9 \\
Cross-attention & 66.4 & 75.7 & 87.5 & 91.4 \\
\bottomrule
\end{tabular}
\label{tab:text_feature_fusion}
\end{table}





\begin{figure}
    \centering
    \includegraphics[width=0.8\linewidth]{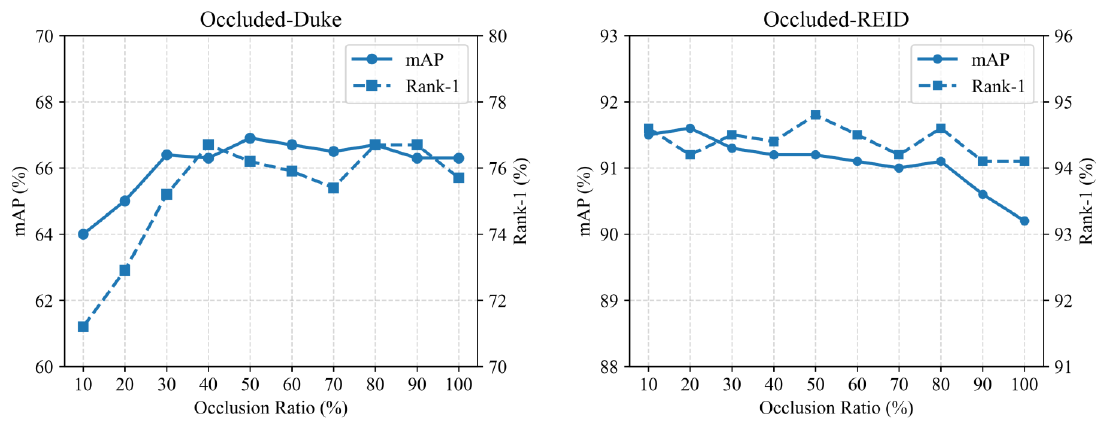}
    \caption{Comparison of occlusion ratios on Occluded-Duke and Occluded-REID.}
    \label{fig:ratio}
\end{figure}

\begin{table}
\centering
\caption{ Comparison of different numbers of learnable tokens M and N for Template$^{occ}$ on Occluded-Duke.}
\begin{tabular}{l c c c c}
\toprule
M,N & mAP & R1 & R5 & R10 \\
\midrule
2,2 & 66.8 & 75.7 & 88.1 & 91.3 \\
2,4 & 66.8 & 75.8 & 87.6 & 91.2 \\
4,2 & 66.3 & 75.0 & \textbf{88.3} & 91.1 \\
4,4 & \textbf{66.9} & \textbf{76.2} & 87.9 & \textbf{91.7} \\
4,6 & 66.5 & 75.8 & 88.0 & 91.0 \\
6,4 & 66.7 & 75.8 & 87.7 & 91.3 \\
6,6 & 66.8 & 75.9 & 88.1 & 91.3 \\
\bottomrule
\end{tabular}
\label{tab:M_N}
\end{table}

\begin{table}
\centering
\caption{Comparison of computational cost and sample throughput (samples/s) at batch size 64 on an NVIDIA A100 GPU.}
\begin{tabular}{l c c c}
\toprule
Method  & Params(M) &  FLOPs(G) & Speed(samples/s) \\
\midrule
CLIP-ReID & 126.68 & 12.07 & 258 \\
DPL-ReID & 142.12 & 12.50 & 173  \\
\bottomrule
\end{tabular}
\label{tab:cost}
\end{table}

\textbf{Impact of occlusion ratio.} 
We analyze the impact of different occlusion ratios in Figure~\ref{fig:ratio}. For Occluded-Duke, as the occlusion ratio increases, the performance gradually improves and eventually stabilizes. In contrast, for Occluded-REID, which is trained using the Market-1501 dataset, there already exist a certain proportion of occluded samples. As the occlusion ratio increases further, it leads to a significant loss of pedestrian information, preventing the model from learning identity-related properties. Therefore, based on the results on these two datasets, we set 0.5 as the default occlusion ratio.

\textbf{Impact of the number of learnable tokens N and M.} 
We follow CLIP-ReID and set the number of learnable tokens in Template$^{full}$ to M=4. We further investigate the effects of different numbers of M and N in Template$^{occ}$ on the Occluded-Duke.
As shown in Table~\ref{tab:M_N}, we finally select N = 4 for X tokens and M = 4 for Y tokens.

\textbf{Computational Cost} The additional computational overhead of our method mainly comes from the WGFF module. Although LSNet is introduced, we adopt a relatively lightweight variant, LSNet-T, compared to common backbones such as ResNet-50 and ViT-B/16, with only 11.4M parameters and 0.3G FLOPs. As shown in Table~\ref{tab:cost}, this module introduces only limited increases in both parameters and computational cost.
Despite the slight increase in training overhead, WGFF effectively alleviates the local attention bias of CNN- and ViT-based feature extractors under occlusion, leading to consistent performance improvements. Therefore, the overhead is well controlled with a favorable trade-off between efficiency and performance.

\begin{figure}
    \centering
    \includegraphics[width=0.5\linewidth]{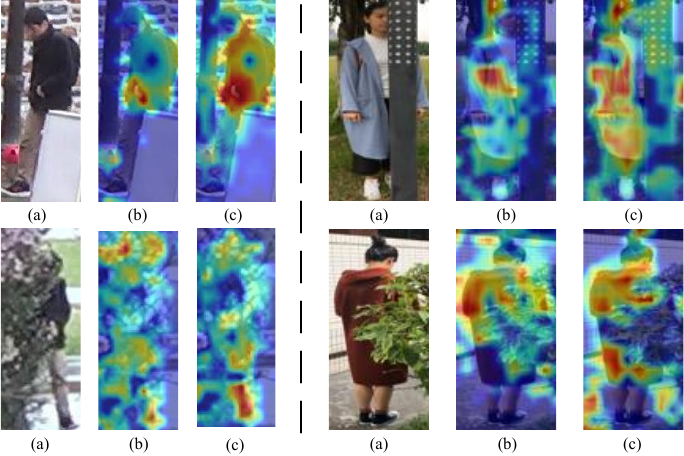}
    \caption{Visualization of ViT-based ReID.
    (a) original image, (b) CLIP-ReID, and (c) DPL-ReID. The left and right images are respectively test samples from Occluded-Duke and Occluded-REID.}
    \label{fig:Visualization}
\end{figure}

\begin{figure}
    \centering
    \includegraphics[width=0.7\linewidth]{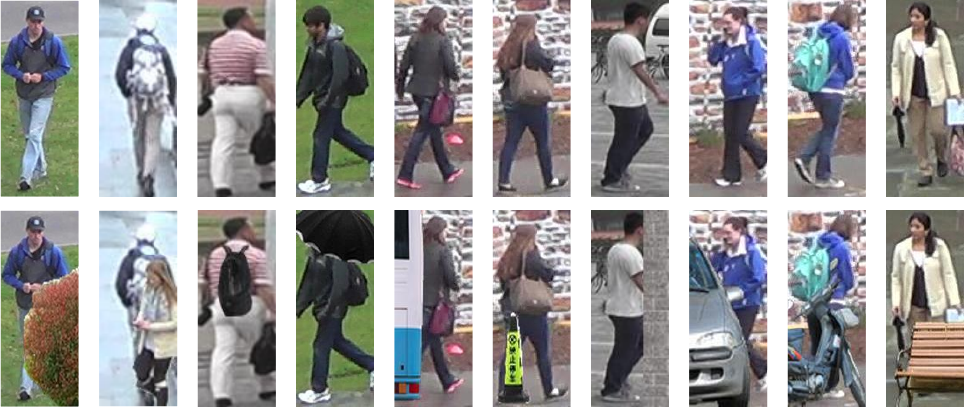}
    \caption{   Visualization of occlusion effects. The top row shows the original images, while the bottom row shows the occlusion images generated by our proposed RWOA.}
    \label{fig:occlusion_effects}
\end{figure}

\begin{figure}
    \centering
    \includegraphics[width=0.8\linewidth]{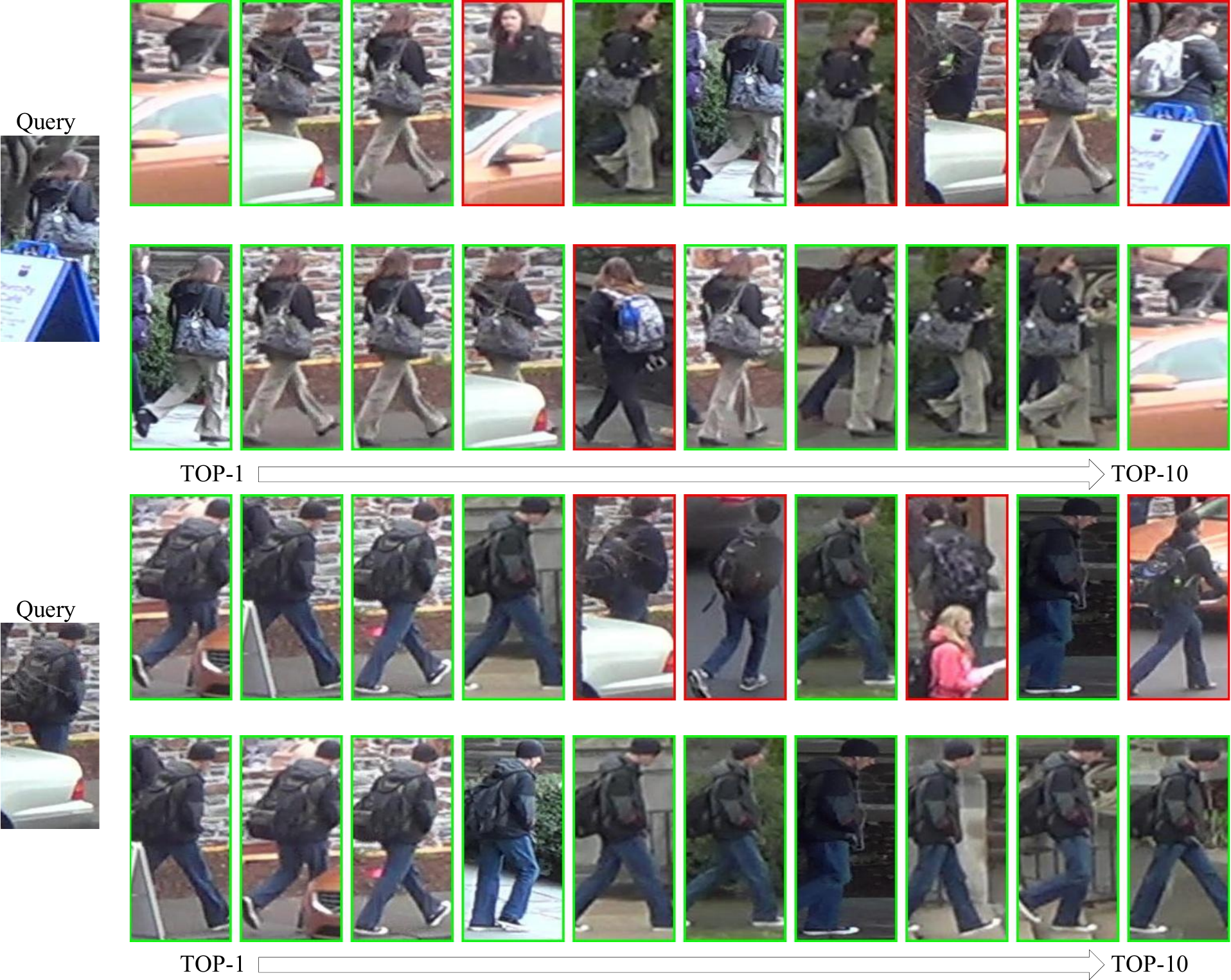}
    \caption{   Retrieval results of baseline CLIP-ReID (the first row) and ours DPL-ReID (the second row). The top row shows the baseline CLIP-ReID, while the bottom row shows DPL-ReID. The incorrectly identified samples are highlighted in red, while the green color highlights the correct samples.}
    \label{fig:querygallery}
\end{figure}

\subsection{Visualization}
\textbf{Attention map visualization.} We employ Grad-CAM to visualize the attention heatmaps of the visual encoder, shown in the Figure~\ref{fig:Visualization}. DPL-ReID focuses on more pedestrian regions compared to CLIP-ReID, due to the combined effect of Dual-PL and WGFF.

\textbf{Visualization of occlusion effects.} Figure~\ref{fig:occlusion_effects} gives some instances of occlusion samples generated by our proposed Real-World Occlusion Augmentation (RWOA) method. Seeing from it, we observed that these generating occlusion images are very close to ones existing in the real world.

\textbf{Retrieval Performance.} We compare the Top-10 retrieval performance of the ViT-based DPL-ReID with the baseline CLIP-ReID on the Occluded-Duke dataset, as shown in Figure~\ref{fig:querygallery}. Our DPL-ReID achieves higher person ReID accuracy than CLIP-ReID.

\section{Conclusion}
In this work, we revisit the role of occluders in occluded person ReID and propose a novel method, DPL-ReID. We also construct an occlusion instance library containing richer and more realistic occluders, making the simulated occlusion closer to real-world scenarios. Additionally, our WGFF effectively addresses the limited receptive field of the model. Both the overall model and individual modules achieve strong performance on occluded ReID datasets.
\section{Acknowledgement}
This work is supported by the National Natural Science Foundation of China (No.61602288, 62676160), Fundamental Research Program of Shanxi Province (No.202503021211073). The authors also would like to thank the anonymous reviewers for their valuable suggestions.

\printcredits

\bibliographystyle{cas-model2-names}

\bibliography{cas-refs}

@article{zheng2015person,
  title={Person re-identification meets image search},
  author={Zheng, Liang and Shen, Liyue and Tian, Lu and Wang, Shengjin and Bu, Jiahao and Tian, Qi},
  journal={arXiv preprint arXiv:1502.02171},
  year={2015}
}

@inproceedings{radford2021learning,
  title={Learning transferable visual models from natural language supervision},
  author={Radford, Alec and Kim, Jong Wook and Hallacy, Chris and Ramesh, Aditya and Goh, Gabriel and Agarwal, Sandhini and Sastry, Girish and Askell, Amanda and Mishkin, Pamela and Clark, Jack and others},
  booktitle={International conference on machine learning},
  pages={8748--8763},
  year={2021},
  organization={PmLR}
}

@inproceedings{wang2025lsnet,
  title={LSNet: See Large, Focus Small},
  author={Wang, Ao and Chen, Hui and Lin, Zijia and Han, Jungong and Ding, Guiguang},
  booktitle={Proceedings of the Computer Vision and Pattern Recognition Conference},
  pages={9718--9729},
  year={2025}
}

@article{zhou2022learning,
  title={Learning to prompt for vision-language models},
  author={Zhou, Kaiyang and Yang, Jingkang and Loy, Chen Change and Liu, Ziwei},
  journal={International Journal of Computer Vision},
  volume={130},
  number={9},
  pages={2337--2348},
  year={2022},
  publisher={Springer}
}

@inproceedings{li2023clip,
  title={Clip-reid: exploiting vision-language model for image re-identification without concrete text labels},
  author={Li, Siyuan and Sun, Li and Li, Qingli},
  booktitle={Proceedings of the AAAI conference on artificial intelligence},
  volume={37},
  number={1},
  pages={1405--1413},
  year={2023}
}

@inproceedings{sun2018beyond,
  title={Beyond part models: Person retrieval with refined part pooling (and a strong convolutional baseline)},
  author={Sun, Yifan and Zheng, Liang and Yang, Yi and Tian, Qi and Wang, Shengjin},
  booktitle={Proceedings of the European conference on computer vision (ECCV)},
  pages={480--496},
  year={2018}
}

@inproceedings{wang2018learning,
  title={Learning discriminative features with multiple granularities for person re-identification},
  author={Wang, Guanshuo and Yuan, Yufeng and Chen, Xiong and Li, Jiwei and Zhou, Xi},
  booktitle={Proceedings of the 26th ACM international conference on Multimedia},
  pages={274--282},
  year={2018}
}

@inproceedings{wang2020high,
  title={High-order information matters: Learning relation and topology for occluded person re-identification},
  author={Wang, Guan'an and Yang, Shuo and Liu, Huanyu and Wang, Zhicheng and Yang, Yang and Wang, Shuliang and Yu, Gang and Zhou, Erjin and Sun, Jian},
  booktitle={Proceedings of the IEEE/CVF conference on computer vision and pattern recognition},
  pages={6449--6458},
  year={2020}
}

@inproceedings{wang2022pose,
  title={Pose-guided feature disentangling for occluded person re-identification based on transformer},
  author={Wang, Tao and Liu, Hong and Song, Pinhao and Guo, Tianyu and Shi, Wei},
  booktitle={Proceedings of the AAAI conference on artificial intelligence},
  volume={36},
  number={3},
  pages={2540--2549},
  year={2022}
}

@inproceedings{sun2019deep,
  title={Deep high-resolution representation learning for human pose estimation},
  author={Sun, Ke and Xiao, Bin and Liu, Dong and Wang, Jingdong},
  booktitle={Proceedings of the IEEE/CVF conference on computer vision and pattern recognition},
  pages={5693--5703},
  year={2019}
}

@article{yuan2021hrformer,
  title={Hrformer: High-resolution transformer for dense prediction},
  author={Yuan, Yuhui and Fu, Rao and Huang, Lang and Lin, Weihong and Zhang, Chao and Chen, Xilin and Wang, Jingdong},
  journal={arXiv preprint arXiv:2110.09408},
  year={2021}
}

@article{xu2022vitpose,
  title={Vitpose: Simple vision transformer baselines for human pose estimation},
  author={Xu, Yufei and Zhang, Jing and Zhang, Qiming and Tao, Dacheng},
  journal={Advances in neural information processing systems},
  volume={35},
  pages={38571--38584},
  year={2022}
}

@inproceedings{zhong2020random,
  title={Random erasing data augmentation},
  author={Zhong, Zhun and Zheng, Liang and Kang, Guoliang and Li, Shaozi and Yang, Yi},
  booktitle={Proceedings of the AAAI conference on artificial intelligence},
  volume={34},
  number={07},
  pages={13001--13008},
  year={2020}
}

@article{devries2017improved,
  title={Improved regularization of convolutional neural networks with cutout},
  author={DeVries, Terrance and Taylor, Graham W},
  journal={arXiv preprint arXiv:1708.04552},
  year={2017}
}

@inproceedings{kumar2017hide,
  title={Hide-and-seek: Forcing a network to be meticulous for weakly-supervised object and action localization},
  author={Kumar Singh, Krishna and Jae Lee, Yong},
  booktitle={Proceedings of the IEEE international conference on computer vision},
  pages={3524--3533},
  year={2017}
}

@inproceedings{yun2019cutmix,
  title={Cutmix: Regularization strategy to train strong classifiers with localizable features},
  author={Yun, Sangdoo and Han, Dongyoon and Oh, Seong Joon and Chun, Sanghyuk and Choe, Junsuk and Yoo, Youngjoon},
  booktitle={Proceedings of the IEEE/CVF international conference on computer vision},
  pages={6023--6032},
  year={2019}
}

@inproceedings{chen2021occlude,
  title={Occlude them all: Occlusion-aware attention network for occluded person re-id},
  author={Chen, Peixian and Liu, Wenfeng and Dai, Pingyang and Liu, Jianzhuang and Ye, Qixiang and Xu, Mingliang and Chen, Qi’an and Ji, Rongrong},
  booktitle={Proceedings of the IEEE/CVF international conference on computer vision},
  pages={11833--11842},
  year={2021}
}

@article{song2024deep,
  title={A deep hierarchical feature sparse framework for occluded person re-identification},
  author={Song, Yihu and Liu, Shuaishi},
  journal={arXiv preprint arXiv:2401.07469},
  year={2024}
}

@inproceedings{wang2022feature,
  title={Feature erasing and diffusion network for occluded person re-identification},
  author={Wang, Zhikang and Zhu, Feng and Tang, Shixiang and Zhao, Rui and He, Lihuo and Song, Jiangning},
  booktitle={Proceedings of the IEEE/CVF conference on computer vision and pattern recognition},
  pages={4754--4763},
  year={2022}
}

@article{wang2024feature,
  title={Feature completion transformer for occluded person re-identification},
  author={Wang, Tao and Liu, Mengyuan and Liu, Hong and Li, Wenhao and Ban, Miaoju and Guo, Tianyu and Li, Yidi},
  journal={IEEE Transactions on Multimedia},
  volume={26},
  pages={8529--8542},
  year={2024},
  publisher={IEEE}
}

@inproceedings{he2017mask,
  title={Mask r-cnn},
  author={He, Kaiming and Gkioxari, Georgia and Doll{\'a}r, Piotr and Girshick, Ross},
  booktitle={Proceedings of the IEEE international conference on computer vision},
  pages={2961--2969},
  year={2017}
}

@article{he2023region,
  title={Region generation and assessment network for occluded person re-identification},
  author={He, Shuting and Chen, Weihua and Wang, Kai and Luo, Hao and Wang, Fan and Jiang, Wei and Ding, Henghui},
  journal={IEEE transactions on information forensics and security},
  volume={19},
  pages={120--132},
  year={2023},
  publisher={IEEE}
}

@article{zhi2025attribute,
  title={Attribute Guidance With Inherent Pseudo-label For Occluded Person Re-identification},
  author={Zhi, Rui and Yang, Zhen and Zhang, Haiyang},
  journal={arXiv preprint arXiv:2508.04998},
  year={2025}
}

@article{huang2025background,
  title={Background Matters Too: A Language-Enhanced Adversarial Framework for Person Re-Identification},
  author={Huang, Kaicong and Azfar, Talha and Reilly, Jack M and Guggisberg, Thomas and Ke, Ruimin},
  journal={arXiv preprint arXiv:2509.03032},
  year={2025}
}

@inproceedings{zheng2017unlabeled,
  title={Unlabeled samples generated by gan improve the person re-identification baseline in vitro},
  author={Zheng, Zhedong and Zheng, Liang and Yang, Yi},
  booktitle={Proceedings of the IEEE international conference on computer vision},
  pages={3754--3762},
  year={2017}
}

@inproceedings{lin2014microsoft,
  title={Microsoft coco: Common objects in context},
  author={Lin, Tsung-Yi and Maire, Michael and Belongie, Serge and Hays, James and Perona, Pietro and Ramanan, Deva and Doll{\'a}r, Piotr and Zitnick, C Lawrence},
  booktitle={European conference on computer vision},
  pages={740--755},
  year={2014},
  organization={Springer}
}

@inproceedings{yu2020bdd100k,
  title={Bdd100k: A diverse driving dataset for heterogeneous multitask learning},
  author={Yu, Fisher and Chen, Haofeng and Wang, Xin and Xian, Wenqi and Chen, Yingying and Liu, Fangchen and Madhavan, Vashisht and Darrell, Trevor},
  booktitle={Proceedings of the IEEE/CVF conference on computer vision and pattern recognition},
  pages={2636--2645},
  year={2020}
}

@inproceedings{he2016deep,
  title={Deep residual learning for image recognition},
  author={He, Kaiming and Zhang, Xiangyu and Ren, Shaoqing and Sun, Jian},
  booktitle={Proceedings of the IEEE conference on computer vision and pattern recognition},
  pages={770--778},
  year={2016}
}

@article{dosovitskiy2020image,
  title={An image is worth 16x16 words: Transformers for image recognition at scale},
  author={Dosovitskiy, Alexey},
  journal={arXiv preprint arXiv:2010.11929},
  year={2020}
}

@inproceedings{gao2020pose,
  title={Pose-guided visible part matching for occluded person reid},
  author={Gao, Shang and Wang, Jingya and Lu, Huchuan and Liu, Zimo},
  booktitle={Proceedings of the IEEE/CVF conference on computer vision and pattern recognition},
  pages={11744--11752},
  year={2020}
}

@inproceedings{he2021transreid,
  title={Transreid: Transformer-based object re-identification},
  author={He, Shuting and Luo, Hao and Wang, Pichao and Wang, Fan and Li, Hao and Jiang, Wei},
  booktitle={Proceedings of the IEEE/CVF international conference on computer vision},
  pages={15013--15022},
  year={2021}
}

@inproceedings{li2021diverse,
  title={Diverse part discovery: Occluded person re-identification with part-aware transformer},
  author={Li, Yulin and He, Jianfeng and Zhang, Tianzhu and Liu, Xiang and Zhang, Yongdong and Wu, Feng},
  booktitle={Proceedings of the IEEE/CVF conference on computer vision and pattern recognition},
  pages={2898--2907},
  year={2021}
}

@article{wang2025looking,
  title={Looking Clearer with Text: A Hierarchical Context Blending Network for Occluded Person Re-Identification},
  author={Wang, Changshuo and He, Shuting and Wu, Meiqing and Lam, Siew-Kei and Tiwari, Prayag and Gao, Xingyu},
  journal={IEEE Transactions on Information Forensics and Security},
  year={2025},
  publisher={IEEE}
}

@inproceedings{miao2019pose,
  title={Pose-guided feature alignment for occluded person re-identification},
  author={Miao, Jiaxu and Wu, Yu and Liu, Ping and Ding, Yuhang and Yang, Yi},
  booktitle={Proceedings of the IEEE/CVF international conference on computer vision},
  pages={542--551},
  year={2019}
}

@inproceedings{zhuo2018occluded,
  title={Occluded person re-identification},
  author={Zhuo, Jiaxuan and Chen, Zeyu and Lai, Jianhuang and Wang, Guangcong},
  booktitle={2018 IEEE international conference on multimedia and expo (ICME)},
  pages={1--6},
  year={2018},
  organization={IEEE}
}

@article{han2023spatial,
  title={Spatial complementary and self-repair learning for occluded person re-identification},
  author={Han, Shoudong and Liu, Donghaisheng and Zhang, Ziwen and Ming, Delie},
  journal={Neurocomputing},
  volume={546},
  pages={126360},
  year={2023},
  publisher={Elsevier}
}

@inproceedings{liu2021swin,
  title={Swin transformer: Hierarchical vision transformer using shifted windows},
  author={Liu, Ze and Lin, Yutong and Cao, Yue and Hu, Han and Wei, Yixuan and Zhang, Zheng and Lin, Stephen and Guo, Baining},
  booktitle={Proceedings of the IEEE/CVF international conference on computer vision},
  pages={10012--10022},
  year={2021}
}

@article{russakovsky2015imagenet,
  title={Imagenet large scale visual recognition challenge},
  author={Russakovsky, Olga and Deng, Jia and Su, Hao and Krause, Jonathan and Satheesh, Sanjeev and Ma, Sean and Huang, Zhiheng and Karpathy, Andrej and Khosla, Aditya and Bernstein, Michael and others},
  journal={International journal of computer vision},
  volume={115},
  number={3},
  pages={211--252},
  year={2015},
  publisher={Springer}
}

@inproceedings{fu2021unsupervised,
  title={Unsupervised pre-training for person re-identification},
  author={Fu, Dengpan and Chen, Dongdong and Bao, Jianmin and Yang, Hao and Yuan, Lu and Zhang, Lei and Li, Houqiang and Chen, Dong},
  booktitle={Proceedings of the IEEE/CVF conference on computer vision and pattern recognition},
  pages={14750--14759},
  year={2021}
}

@inproceedings{selvaraju2017grad,
  title={Grad-cam: Visual explanations from deep networks via gradient-based localization},
  author={Selvaraju, Ramprasaath R and Cogswell, Michael and Das, Abhishek and Vedantam, Ramakrishna and Parikh, Devi and Batra, Dhruv},
  booktitle={Proceedings of the IEEE international conference on computer vision},
  pages={618--626},
  year={2017}
}

@inproceedings{jia2021matching,
  title={Matching on sets: Conquer occluded person re-identification without alignment},
  author={Jia, Mengxi and Cheng, Xinhua and Zhai, Yunpeng and Lu, Shijian and Ma, Siwei and Tian, Yonghong and Zhang, Jian},
  booktitle={Proceedings of the AAAI conference on artificial intelligence},
  volume={35},
  number={2},
  pages={1673--1681},
  year={2021}
}

@inproceedings{zheng2021pose,
  title={Pose-guided feature learning with knowledge distillation for occluded person re-identification},
  author={Zheng, Kecheng and Lan, Cuiling and Zeng, Wenjun and Liu, Jiawei and Zhang, Zhizheng and Zha, Zheng-Jun},
  booktitle={Proceedings of the 29th ACM international conference on multimedia},
  pages={4537--4545},
  year={2021}
}

@article{Sun2019MVPMA,
  title={MVP Matching: A Maximum-Value Perfect Matching for Mining Hard Samples, With Application to Person Re-Identification},
  author={Han Sun and Zhiyuan Chen and Shiyang Yan and Lin Xu},
  journal={2019 IEEE/CVF International Conference on Computer Vision (ICCV)},
  year={2019},
  pages={6736-6746},
  url={https://api.semanticscholar.org/CorpusID:204958086}
}

@article{Tay2019AANetAA,
  title={AANet: Attribute Attention Network for Person Re-Identifications},
  author={Chiat-Pin Tay and Sharmili Roy and Kim-Hui Yap},
  journal={2019 IEEE/CVF Conference on Computer Vision and Pattern Recognition (CVPR)},
  year={2019},
  pages={7127-7136},
  url={https://api.semanticscholar.org/CorpusID:195823196}
}

@article{Zhu2021AAformerAT,
  title={AAformer: Auto-Aligned Transformer for Person Re-Identification},
  author={Kuan Zhu and Haiyun Guo and Shiliang Zhang and Yaowei Wang and Gaopan Huang and Honglin Qiao and Jing Liu and Jinqiao Wang and Ming Tang},
  journal={IEEE Transactions on Neural Networks and Learning Systems},
  year={2021},
  volume={35},
  pages={17307-17317},
  url={https://api.semanticscholar.org/CorpusID:233004431}
}

@article{Huang2017ArbitraryST,
  title={Arbitrary Style Transfer in Real-Time with Adaptive Instance Normalization},
  author={Xun Huang and Serge J. Belongie},
  journal={2017 IEEE International Conference on Computer Vision (ICCV)},
  year={2017},
  pages={1510-1519},
  url={https://api.semanticscholar.org/CorpusID:6576859}
}

@article{Ji2025Exploring,
  author    = {Z. Ji and D. Cheng and K. Feng},
  title     = {Exploring stronger transformer representation learning for occluded person re-identification},
  journal   = {Multimedia Systems},
  volume    = {31},
  pages     = {394},
  year      = {2025},
  doi       = {10.1007/s00530-025-01986-0},
  url       = {https://doi.org/10.1007/s00530-025-01986-0}
}

@article{10.1016/j.neucom.2024.127442,
author = {Zhang, Ziwen and Han, Shoudong and Liu, Donghaisheng and Ming, Delie},
title = {Focus and imagine: Occlusion suppression and repairing transformer for occluded person re-identification},
year = {2024},
issue_date = {Apr 2024},
publisher = {Elsevier Science Publishers B. V.},
address = {NLD},
volume = {578},
number = {C},
issn = {0925-2312},
url = {https://doi.org/10.1016/j.neucom.2024.127442},
doi = {10.1016/j.neucom.2024.127442},
journal = {Neurocomput.},
month = apr,
numpages = {11}
}



\end{document}